\documentclass{Interspeech2024}

\interspeechcameraready 
\usepackage{tipa}
\usepackage[utf8]{inputenc}
\usepackage{xcolor}
\usepackage{multirow}
\DeclareUnicodeCharacter{02CC}{\hspace{0pt}}
\DeclareUnicodeCharacter{02C8}{\hspace{0pt}}

\title{Enhancing Out-of-Vocabulary Performance of Indian TTS Systems for Practical Applications through Low-Effort Data Strategies}

\name[]{Srija}{Anand}
\name[]{Praveen}{Srinivasa Varadhan}
\name[]{Ashwin}{Sankar}
\name[]{Giri}{Raju}
\name[]{Mitesh}{M. Khapra}

\address{
  AI4Bharat, Indian Institute of Technology, Madras, India
}
\email{srijaanand@ai4bharat.org, cs21d201@cse.iitm.ac.in, ashwins1211@gmail.com, girirajur023@gmail.com, miteshk@cse.iitm.ac.in}


\newcommand{\dataset}[1]{\textsc{IndicOOV}}

\begin{document}
\maketitle

\begin{abstract}
Publicly available TTS datasets for low-resource languages like Hindi and Tamil typically contain 10-20 hours of data, leading to poor vocabulary coverage. This limitation becomes evident in downstream applications where domain-specific vocabulary coupled with frequent code-mixing with English, results in many OOV words. To highlight this problem, we create a benchmark containing OOV words from several real-world applications. Indeed, state-of-the-art Hindi and Tamil TTS systems perform poorly on this OOV benchmark, as indicated by intelligibility tests. To improve the model’s OOV performance, we propose a low-effort and economically viable strategy to obtain more training data. Specifically, we propose using volunteers as opposed to high quality voice artists to record words containing character bigrams unseen in the training data. We show that using such inexpensive data, the model's performance improves on OOV words, while not affecting voice quality and in-domain performance.
\end{abstract}

\section{Introduction}
\begin{figure}[!t]
    \centering
    \includegraphics[width=\columnwidth]{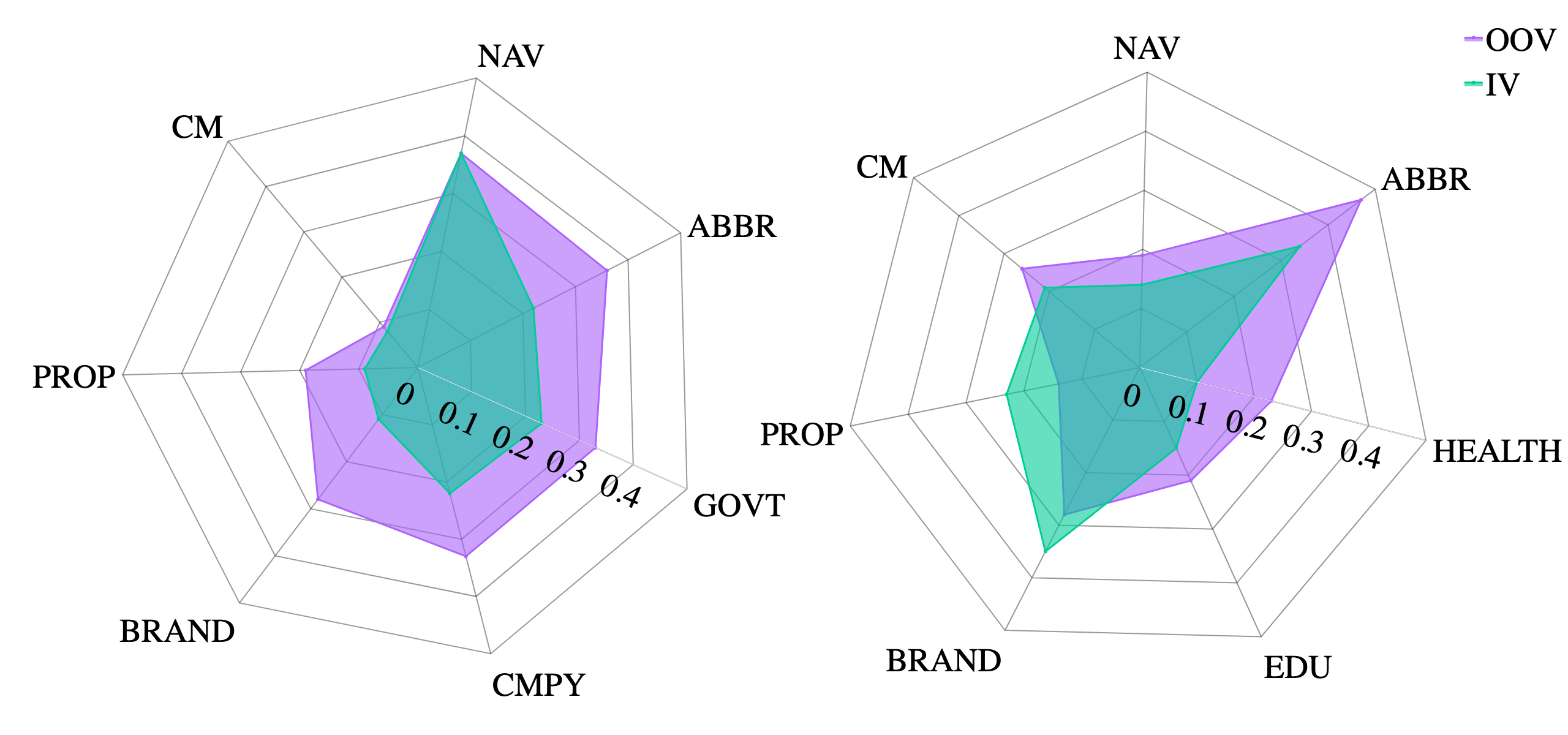}
    \caption{Intelligibility Error Rates (\%) of Indian TTS models across Hindi (Left) and Tamil (Right) on the \dataset{} benchmark shows that models consistently perform worse for OOV words compared to IV words.}
    \label{fig:OOV_id}
\end{figure}

Text-to-Speech (TTS) systems play a crucial role in linguistically diverse and developing regions like India, finding usage in various commercial and governmental applications. For example, they can be used for broadcasting vital information to farmers about weather conditions, disseminating information about government schemes, and enhancing accessibility for the visually impaired. However, the effectiveness of these systems is often hampered by limited training data.  Publicly available TTS datasets for languages such as Hindi and Tamil typically range from 10-20 hours \cite{baby2016resources, srivastava2022uss}, resulting in inadequate vocabulary coverage. This shortfall becomes particularly evident in practical applications, where the occurrence of out-of-vocabulary (OOV) words is inevitable, due to frequent code-mixing with English as well as the usage of specialized domain-specific vocabulary. While this issue is well-documented in English TTS systems \cite{li2023styletts, shen2024naturalspeech, basetts, ju2024naturalspeech}, in this work, we show that this is also the case for low-resource languages like Hindi and Tamil, where TTS systems similarly under-perform on OOV words compared to in-vocabulary (IV) words (see Figure \ref{fig:OOV_id}).

Given the above situation, our goal is to improve the intelligibility of TTS systems on OOV texts while retaining their naturalness. However, several challenges exist when attempting to improve the OOV performance of existing systems. First, it would be ideal to record more training data containing OOV words using the same speaker from the original dataset that the TTS model was trained on because such speakers are typically carefully selected artists with pleasant voices and speaking styles that are more suited for building TTS systems. However, this is mostly infeasible because the identities of the original speakers are anonymized for ethical reasons. 
Second, even if we had access to the original speaker, one would still need access to a larger corpus for low-resource languages to carefully curate a set of OOV words or sentences that can be recorded and later used to improve a model's performance. Finally, one would need a robust benchmark with broad coverage of OOV words across domains to assess whether performance gains generalize well across different real-world applications.

We attempt to tackle all of these challenges in the context of Hindi and Tamil, to address the gap between TTS performance on OOV and IV texts. Since we do not have access to the speakers in the original dataset, we instead explore a cost-effective alternative of recording OOV words from volunteers having different voices \cite{abraham2020crowdsourcing} and evaluate whether training the TTS system on this new data can reduce the OOV intelligibility error rates for the original voice of the TTS system. Next, to curate OOV texts for recording, we collate an extensive corpus from multiple resources for Hindi and Tamil and then carefully select OOV words by maximizing the coverage of high-frequency missing OOV character bigrams. 
This is motivated by the importance of achieving syllabic balance \cite{ai4bharat2024rasa} and the importance of phonotactics \cite{prakash2023exploring} in TTS for Indian languages. 

Finally, to evaluate the performance gains of our proposed approach, we release a benchmark, \dataset{}, for Hindi and Tamil containing OOV words which are not seen in the original training data as well as in the inexpensive data recorded using volunteers. These OOV words span 7 categories, viz. Abbreviations, Brands and Products, Codemixed (English-Hindi, English-Tamil), Company Names, Government Schemes, Proper Nouns and Navigations, which are typically seen in downstream applications. \dataset{} contains 100 sentences per category for a comprehensive evaluation. We conduct intelligibility tests and show that our cost-effective method indeed leads to better performance on \dataset{}, while not affecting the voice quality obtained by just training on the original speaker's data. 

\section{\dataset{} Benchmark}
\label{sec:benchmark}

\noindent We present \dataset{}\footnote{GitHub repository for IndicOOV: \url{https://github.com/AI4Bharat/IndicOOV/}}, a novel benchmark designed to evaluate the out-of-vocabulary (OOV) word synthesis capabilities of Text-to-speech (TTS) systems for two Indian languages - Hindi and Tamil. A key challenge in creating a benchmark across categories covering real-world applications is the lack of a readily available corpus in Indian languages containing labeled texts covering different categories.  We thus break down the process of creating the benchmark into three steps - (i) Find in-vocabulary (IV) and OOV words from a larger text corpus, (ii) classify found words into categories, and (iii) filter out unsuitable words.

We first collate a larger corpus of texts across Hindi and Tamil from diverse datasets. We primarily rely on texts from Sangraha \cite{khan2024indicllmsuite}, the Bharat Parallel Corpus \cite{gala2023indictrans2}, and transcriptions from IndicVoices \cite{javed2024indicvoices}. All these corpora together help us gather words from a variety of sources like Wikipedia, Pratham Books, National Institute of Open Schooling, Press Information Bureau, Mann Ki Baat, and other government and open websites, that reflect practical scenarios one would potentially deploy TTS systems in.

We then programmatically identify words containing high-frequency OOV character bigrams missing in the IndicTTS \cite{baby2016resources} training corpus. We form a set of over 1000 words and task a human language expert to scan this word list and manually classify individual words into the 7 categories of interest mentioned earlier. Once we reached fifty words within a particular category, we requested the language expert to prioritize finding words in other categories where the target counts weren't met. For some categories, such as Proper nouns or Navigation phrases, even with multiple iterations, we could not find the desired number of words containing OOV character bigrams. In such cases, we created a set of missing character OOV bigrams and relied on the creativity of the language expert to create or recollect words for a particular category containing these OOV bigrams. Likewise, we also curate a list of fifty IV words, which have all their bigrams present in the IndicTTS dataset, for each category.

In this manner, we are able to create a diverse benchmark spanning 7 key TTS application categories: Abbreviations (Abbr), Brands and Products (Brand), Codemixed (CM), Company Names (Cmpy), Government Schemes (Govt), Proper Nouns (Prop), and Navigation (Nav). For Tamil, we replace Company Names and Government Schemes with Education(Edu) and Healthcare(Health) due to the difficulty in obtaining OOV words, respectively.

\section{Recording OOV Words with Volunteers}
\label{section:Methodologies}

\noindent We explore a cost-effective method to improve TTS performance by recording OOV words with the help of volunteers who are not professional voice artists. To do this we require (i) recording scripts, (ii) volunteers willing to lend their voice, and (iii) a recording setup and process to record all data.

\begin{table}[!t]
\fontsize{8pt}{8pt}\selectfont 
\centering
\begingroup

\caption{Statistics of recorded data for the three speakers: Male-1 (M1), Male-2 (M2) and Female-1(F1). The duration is reported in minutes (MM:SS)}
\label{tab:ood-recordings-data}
\setlength{\tabcolsep}{3pt} 
\renewcommand{\arraystretch}{1} 
\begin{tabular}{@{}lcccccc@{}}
\toprule

 Speaker & \multicolumn{3}{c}{Hindi} & \multicolumn{3}{c}{Tamil} \\
 \cmidrule(l){2-4} \cmidrule(l){5-7}
 & \multicolumn{1}{l}{Words} & \multicolumn{1}{l}{Duration} & \multicolumn{1}{l}{Bigrams} & \multicolumn{1}{l}{Words} & \multicolumn{1}{l}{Duration} & \multicolumn{1}{l}{Bigrams} \\ \midrule
M1 & 372 & 08:40 & 2439 & 476 & 14:13 & 3930 \\
M2 & 621 & 15:40 & 4978 & 547 & 21:37 & 4323 \\
F1 & 990 & 23:46 & 7973 & 1075 & 28:41 & 8289 \\ \midrule
Total & 1983 & 48:07 & 15390 & 2098 & 64:32 & 16542 \\ \bottomrule
\end{tabular}
\endgroup
\label{tab:recording statistics}
\end{table}

\subsection{Recording Script Creation}

While creating recording scripts we focus on maximizing the coverage of OOV words along with missing OOV bigrams present in them. Instead of recording semantically meaningful sentences which are complete, we prepare a list of OOV words and randomly join sets of five words separated by commas to create unique utterances. We ensure no word repeats across utterances. To select the OOV words, we iterate through the text corpora collated in Section \ref{sec:benchmark} and use a greedy algorithm to find words that maximize the frequency of missing OOV bigrams in the selected set. Owing to limitations in recording capacity and budget, we restrict our recordings to approximately 2000 words per language. In light of this restriction, we adjust the greedy algorithm to exclude a character bi-gram from consideration once its frequency in the chosen set surpasses $k$. We empirically choose $k=6$, and this parameter can be raised when recording additional data is necessary. Note that none of the OOV words present in \dataset{}, are allowed to be a part of our recording scripts (although OOV words in our recording scripts may share bigrams with OOV words in \dataset{}.)

\subsection{Gathering Volunteers}
We found interested volunteers by circulating a form within our institution. Prior to recording, all volunteers were clearly informed regarding the intended use of their voice data and the compensation they would receive for the same. Following this explanation, speakers were presented with consent forms, which they reviewed and signed.  These forms explicitly documented their informed consent for the utilization of their recordings in the training of speech synthesis systems. To safeguard speaker privacy and prevent potential misuse of the speech data, we have opted to withhold the recorded audio samples from public release.  However, we release the recording scripts employed during data collection to facilitate reproducibility and transparency in research methodology. The entire process of data collection was approved by our Institute Ethics Committee with the compensation in line with recommended norms. 

\subsection{Recording Setup and Process}

Since renting a professional studio can be expensive, we instead rely on recording data in acoustic pods present in our workspace. We record audio using a professional condenser microphone with a pop filter to record the volunteers' voices. Getting the words to be spoken out clearly was crucial for our experiments. To ensure clarity in voice, we ascertained that participants were well-hydrated before recording sessions. A few volunteers indicated that the OOV words were difficult to read. We thus requested all volunteers to practice with the recording script before recording sessions. During the recording, an expert proficient in the language listened to all recordings live and pointed out any pronunciation mistakes made by the speaker. During the recording process, this expert also filtered out words that were deemed inappropriate due to spelling mistakes, archaic nuances, profanity, or toxicity.
\noindent In this manner,  we record a total of 6 speakers, consisting of two male speakers and one female speaker for Tamil and Hindi each. The detailed statistics of the recordings are present in Table \ref{tab:ood-recordings-data}. Note that the use of volunteers as opposed to professional voice artists made, ensured that the data collection was relatively inexpensive with an 85\% reduction in costs and the entire process was completed in 2 working days (counting studio time and post-processing time).

\section{Experimental Setup}
\subsection{Dataset:} 
We use the IndicTTS dataset \cite{baby2016resources} for all experiments. Specifically, we train models on the Hindi and Tamil subsets. The Hindi subset contains approximately 10 hours and 4 minutes of female speech and 10 hours and 5 minutes of male speech. Similarly, the Tamil subset contains approximately 10 hours and 2 minutes of female speech and 10 hours and 33 minutes of male speech. Additionally, we finetune models on the recorded data described in Section \ref{section:Methodologies}.

\subsection{Models} 
We train and evaluate with two state-of-the-art text-to-speech (TTS) models: FastPitch (FP)\cite{lancucki2021fastpitch} and VITS \cite{kim2021conditional}.
We fine-tune a pre-trained FastPitch \cite{Kumar2022Towards}, a non-autoregressive transformer-based spectrogram prediction model, starting from the open-sourced pre-trained checkpoint on IndicTTS. To learn durations the model employs an unsupervised alignment learning framework \cite{badlani2022one} that aligns textual features with acoustic representations.  Mel-spectrogram outputs are then converted to audio waveforms using the HiFiGAN V1 vocoder \cite{kong2020hifi}, pre-trained on the IndicTTS corpus from a publicly available checkpoint \cite{Kumar2022Towards}.  We fine-tune this vocoder on our internal dataset for better speaker generalization. We also train VITS, an end-to-end speech synthesis model, with the same hyperparameter settings as prior work \cite{Kumar2022Towards}. Before training, all the audio samples were downsampled to \SI{22050}{\hertz} and converted to a mono-channel configuration. All the models were trained on an NVIDIA A100 40GB GPU, with a batch size of 16, upto 2500 epochs or until convergence.

\subsection{Evaluation Metrics}
We measure both the intelligibility and perceptual quality of the speech generated by the models. We rely on human intelligibility tests to assess the intelligibility of TTS systems on OOV and IV words. In this test, a rater proficient in the language is tasked to listen to an audio sample along with the complete corresponding text and the benchmark word-of-interest highlighted. The rater is then asked to provide a binary rating of whether the word-of-interest was intelligible or not. Raters are instructed to penalize partially intelligible words too, and mark them as ``not intelligible''.  Raters are provided options for slowing down the audio and selecting and playing segments in repeat if required. Furthermore, raters were encouraged to discuss with each other in case of confusion. All raters who participated in the test were expert listeners with prior experience in evaluating TTS systems and who also aid in quality assurance of high-quality TTS data collection efforts. In our evaluations, we rely on 8 such expert listeners and report the percentage of unintelligible words as the Intelligibility Error Rate (\%). 
Next, to evaluate the perceptual quality of generated speech we rely on two objective metrics - (i) VISQOL \cite{hines2015visqol, chinen2020visqol} and (ii) S-SIM or speaker similarity. VISQOL is a perceptual speech quality estimator that uses a spectro-temporal measure to measure the similarity between ground-truth and reference speech. We use this measure to evaluate the retention of speech quality when training with additional OOV data recorded with low expenses. Since we record data in multiple voices from volunteers and use it to attempt to improve the OOV performance of the original TTS speaker, we measure the speaker similarity of the synthesized model outputs with that of the original speaker.  To compute speaker similarity we report the cosine-similarity between the ground-truth and synthesized samples using embeddings extracted from Titanet \cite{koluguri2022titanet}.
\section{Results}

\subsection{Comparison of TTS on IV v/s OOV Texts}
\begin{figure}[!t]
    \centering    \includegraphics[width=\columnwidth]{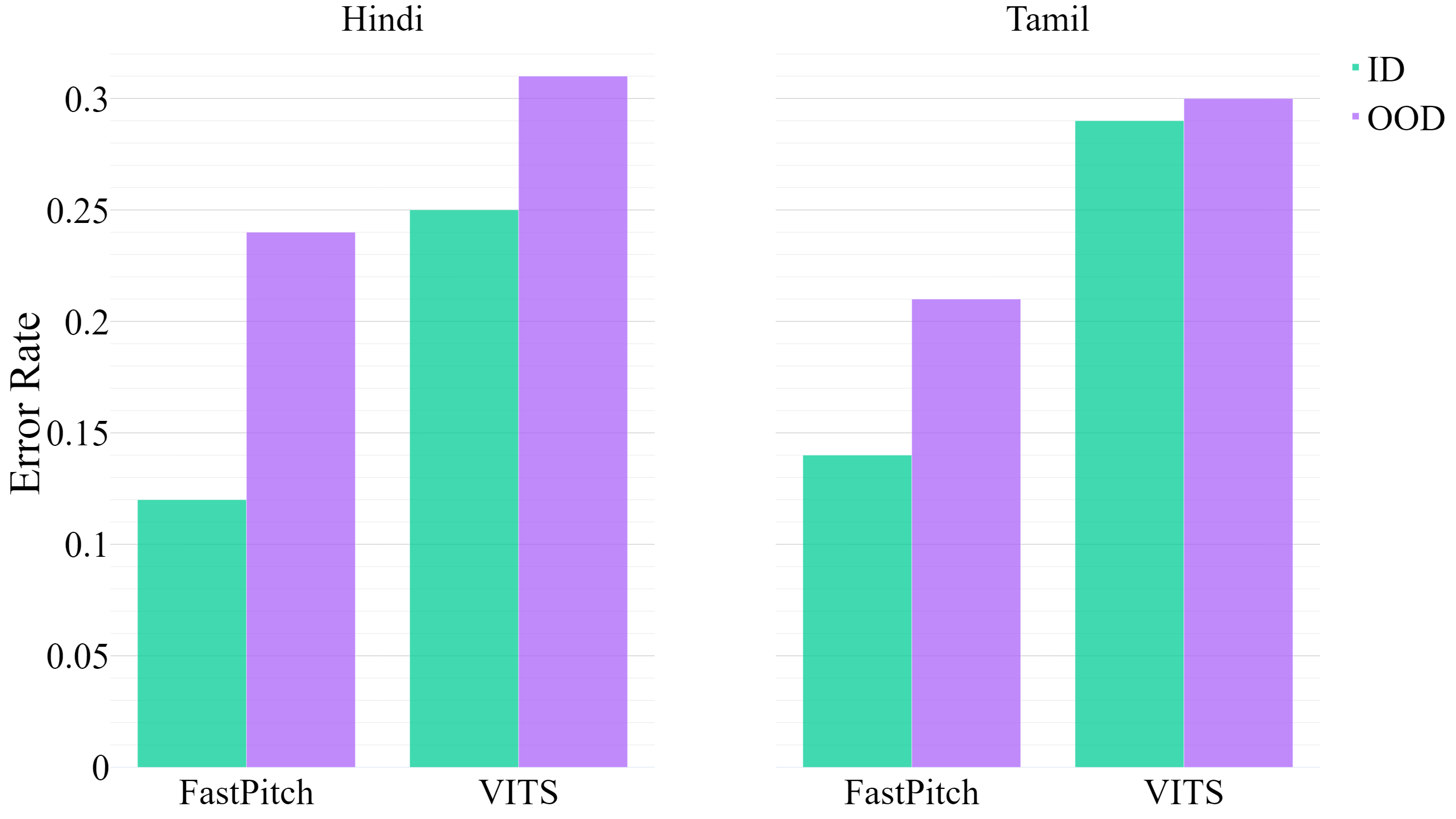}
    \caption{Intelligibility Error Rates (\%) of Indian TTS models across Hindi (Left) and Tamil (Right) on the \dataset{} benchmark averaged across categories shows that models consistently perform worse for OOV words compared to IV words.}
    \label{fig:id_vs_ood_avg}
\end{figure}

In Figure \ref{fig:id_vs_ood_avg}, we visualize the intelligibility error rates of the four baseline models - FastPitch and VITS, each trained on Hindi and Tamil data. Clearly, all models perform worse on OOV words in comparison to IV words across categories. The most noticeable difference is in the Hindi FastPitch, showing high intelligibility error rates of 24\% on OOV words which is twice the IV error rate at 12\%. Surprisingly, all models show relatively high intelligibility error rates on IV words, too, but this possibly reflects on the low-resource setups on which these systems have been trained.

\subsection{Recording OOV Words Improves OOV Performance}

In Table \ref{tab:benchmark-main-results}, we compare the intelligibility of baseline models and models fine-tuned on OOV recordings. On average, across categories, the intelligibility error rates of the fine-tuned models reduce in comparison to the baselines for both OOV and IV words.
More specifically, the overall relative reduction in OOV errors is 40.59\%, 36.33 \%, 22.50 \%, and 30.94\% for Hindi FP, Tamil FP, Hindi VITS, and Tamil VITS respectively. This clearly shows the utility of low-cost recordings in significantly improving model performance on OOV (as well as IV) words. 

\begin{table}[h]
\fontsize{8pt}{8pt}\selectfont
\centering
\begingroup
\caption{Intelligibility Error Rates (\%) conducted for baseline models and models fine-tuned on OOV recordings for both In-Vocabulary (I) and OOV (O) test sets, averaged across male and female TTS voices.}
\label{tab:benchmark-main-results}
\setlength{\tabcolsep}{1pt} 
\renewcommand{\arraystretch}{1} 
\begin{tabular}{@{}lcccccccccc@{}}
\toprule
Lang. & Sys. & Train & Test & \multicolumn{1}{l}{Abbr} & \multicolumn{1}{l}{Brand} & \multicolumn{1}{l}{CM} & \multicolumn{1}{l}{{\begin{tabular}[c]{@{}c@{}}CMPY/ \\ Edu\end{tabular}}} & \multicolumn{1}{l}{{\begin{tabular}[c]{@{}c@{}}Govt/ \\ Health\end{tabular}}} & \multicolumn{1}{l}{Prop} & \multicolumn{1}{l}{Nav} \\ \midrule
Hindi & FP & I & I & 0.15 & 0.12 & 0.01 & 0.09 & 0.17 & 0.09 & 0.22 \\
 &  &  & O & 0.35 & 0.27 & 0.08 & 0.31 & 0.28 & 0.21 & 0.21 \\
 &  & I + O & I & 0.13 & 0.11 & 0.03 & 0.03 & 0.12 & 0.08 & 0.15 \\
 &  &  & O & \textbf{0.22} & \textbf{0.10} & \textbf{0.03} & \textbf{0.19} & \textbf{0.21} & \textbf{0.11} & \textbf{0.16} \\
 \cmidrule{2-11}
 & VITS & I & I & 0.29 & 0.09 & 0.15 & 0.34 & 0.28 & 0.09 & 0.51 \\
 &  &  & O & 0.37 & 0.30 & 0.10 & 0.35 & 0.38 & 0.18 & 0.52 \\
 &  & I + O & I & 0.27 & 0.11 & 0.12 & 0.23 & 0.29 & 0.05 & 0.45 \\
 &  &  & O & \textbf{0.23} & \textbf{0.22} & \textbf{0.07} & \textbf{0.25} & 0.38 & \textbf{0.16} & \textbf{0.41} \\
\midrule
Tamil & FP & I & I & 0.21 & 0.32 & 0.15 & 0.07 & 0.06 & 0.09 & 0.11 \\
 &  &  & O & 0.42 & 0.21 & 0.10 & 0.18 & 0.28 & 0.10 & 0.17 \\
 &  &  I + O & I & 0.29 & 0.03 & 0.03 & 0.02 & 0.01 & 0.06 & 0.04 \\
 &  &  & O & 0.47 & \textbf{0.07} & \textbf{0.08} & \textbf{0.05} & \textbf{0.06} & 0.12 & \textbf{0.08} \\
 \cmidrule{2-11}
 & VITS & I & I & 0.48 & 0.39 & 0.26 & 0.22 & 0.13 & 0.37 & 0.18 \\
 &  &  & O & 0.51 & 0.35 & 0.41 & 0.23 & 0.19 & 0.19 & 0.21 \\
 &  & I + O & I & 0.33 & 0.23 & 0.13 & 0.11 & 0.09 & 0.24 & 0.10 \\
 &  &  & O & \textbf{0.47} & \textbf{0.16} & \textbf{0.29} & \textbf{0.15} & \textbf{0.09} & 0.20 & \textbf{0.09} \\ \bottomrule
\end{tabular}
\endgroup
\end{table}

\subsection{Quality of Synthesis}
To assess whether the quality of the TTS output in the original speaker's voice degrades when adding training data from alternative amateur speakers, we compare the voice quality of the fine-tuned models and baselines with two metrics, viz., ViSQOL for perceptual quality estimation and S-SIM for speaker similarity. 
Both S-SIM and VISQOL are full-reference metrics and require ground truth samples to be provided as reference audios. To assess the relative quality of audios with respect to the original IndicTTS speakers, we use the test set utterances as references. 
We first compute VISQOL and S-SIM of the baseline (Base) with respect to the reference. We then compute VISQOL and S-SIM of the fine-tuned model (I+O) with respect to the reference. We observe that the speaker similarity for baseline, S-SIM (Base) and the speaker similarity for the fine-tuned model, S-SIM (I+O) are comparable across languages, models, and TTS voice, indicating that training on OOV recordings of alternate speakers does not degrade the voice of the original TTS speaker. Likewise, the VISQOL scores of the baseline method, VISQOL (Base) and the scores for the fine-tuned model VISQOL (I + O) are comparable indicating there is no degradation in the perceptually quality of speech too. 

\begin{table}[h]
\fontsize{8pt}{8pt}\selectfont
\centering
\caption{Objective Evaluation of TTS models for speaker similarity and perceptual quality. Here, Base refers to the baseline system and I + O are models finetuned on the OOV Recordings. The TTS Voice is the voice of the speakers of IndicTTS samples.}
\begingroup
\setlength{\tabcolsep}{3pt} 
\renewcommand{\arraystretch}{1} 
\begin{tabular}{@{}lllcccc@{}}
\toprule
Lang. & Model & TTS voice & \multicolumn{1}{l}{{\begin{tabular}[c]{@{}c@{}}S-SIM \\ (Base)\end{tabular}}} & \multicolumn{1}{l}{{\begin{tabular}[c]{@{}c@{}}S-SIM \\ (I + O)\end{tabular}}}  & \multicolumn{1}{l}{{\begin{tabular}[c]{@{}c@{}}VISQOL\\ (Base)\end{tabular}}} & \multicolumn{1}{l}{{\begin{tabular}[c]{@{}c@{}}VISQOL\\ (I + O)\end{tabular}}} \\ \midrule
Hindi & VITS & Female & 0.87 & 0.85 & 3.12 & 3.06 \\
 &  & Male & 0.87 & 0.87 & 3.45 & 3.48 \\ \cmidrule(r){2-7} 
 & FastPitch & Female & 0.80 & 0.79 & 3.02 & 3.09 \\
 &  & Male & 0.79 & 0.78 & 3.44 & 3.44 \\ \midrule
Tamil & VITS & Female & 0.83 & 0.84 & 2.81 & 2.95 \\
 &  & Male & 0.82 & 0.83 & 2.94 & 2.99 \\ \cmidrule(r){2-7} 
 & FastPitch & Female & 0.74 & 0.71 & 2.95 & 2.87 \\
 &  & Male & 0.71 & 0.70 & 2.91 & 2.93 \\ \bottomrule
\end{tabular}
\endgroup
\label{tab:voice-quality}
\end{table}

\subsection{Can only single-gender data improve multi-speaker TTS OOV performance?}
We finetune the FastPitch model with single-gender recordings - (i) Using only one female speaker - F1  and (ii) Using only two male speakers - M1 and M2 . We aim to check if improvements in the intelligibility are agnostic to the gender of the recording volunteer. Table \ref{tab:single-speaker-ft} summarizes the intelligibility scores averaged across seven categories for the base FastPitch model and models fine-tuned on male (Base + M1 + M2) and female (Base + F1) speakers respectively. 
We observe from the scores that fine-tuning the model with data from one gender improves the OOV performance across all TTS voices (M \& F) except for the Tamil male voice when fine-tuning only on Male gender data. Furthermore, fine-tuning the baseline model on only Female gender data reduces the intelligibility error rates from 0.28 to 0.15 for the female speaker and 0.20 to 0.11 for the male speaker in Hindi. Similiarly, fine-tuning the baseline model on only Male gender data reduces the intelligibility error rates from 0.28 to 0.12 for the female Hindi speaker and 0.30 to 0.13 for the female Tamil speaker. This shows that one may train a model on single gender OOV recordings and expect to get OOV performance improvements across both male and female TTS output voices.

\begin{table}[h]
\fontsize{8pt}{8pt}\selectfont
\centering
\caption{OOV Errors for FastPitch models finetuned with recordings of only Male vs only Female voices.}
\begingroup
\setlength{\tabcolsep}{3pt} 
\renewcommand{\arraystretch}{1} 
\begin{tabular}{@{}lcccccc@{}}
\toprule
Lang. & \multicolumn{2}{c}{Base} & \multicolumn{2}{c}{Base + M1 + M2} & \multicolumn{2}{c}{Base + F1} \\ \cmidrule(lr{0.1pt}){2-3} \cmidrule(lr{0.1pt}){4-5} \cmidrule(lr{0.1pt}){6-7}
 & F & M & F & M & F & M \\\midrule
Hindi & 0.28 & 0.20 & 0.12 & 0.13 & 0.15 & 0.11 \\
Tamil & 0.30 & 0.11 & 0.13 & 0.13 & 0.14 & 0.09 \\ \bottomrule
\end{tabular}
\endgroup
\label{tab:single-speaker-ft}
\end{table}

\subsection{Common pronunciation errors}
In the baseline models, we find some interesting trends in the pronunciation errors for both FastPitch and VITS, across languages. The pronunciations were not sharp for consecutive vowels in Abbreviations like AISEC (\textipa{/ˌa:iˌe:sˈi:si/})and IAEA\textipa({/ˌa:ie:ˈi:e:/}). We find that consecutive vowel combinations are not common in the IndicTTS dataset, with only 278 words and 12 words with such combinations from 188K and 100K word corpus for Hindi and Tamil respectively. In contrast, the English TTS dataset LJSpeech \cite{ljspeech17} has 36.5K combinations of consecutive vowels in a corpus of 212K words. Such combinations are prevalent in categories like Abbreviations, Government Schemes, and Navigation that borrow English words.\\

\section{Conclusion}
We study the problem of OOV words in practical deployments of TTS systems. We first create a benchmark, \dataset{}, for assessing the performance of TTS systems for Hindi and Tamil. We then show that there is indeed a clear gap in the performance of state-of-the-art TTS systems on IV v/s OOV words as evaluated on \dataset{}. We then propose a low-cost approach for augmenting existing TTS datasets with recordings of OOV texts using amateur voice artists. Finally, we show that training a TTS system of such OOV recordings indeed improves the performance on \dataset{} while not affecting the voice quality of the synthesized outputs.


\section{Acknowledgements}
This project was made possible through the dedicated efforts and collaboration of numerous organizations and participants. We extend our gratitude to Digital India Bhashini, the Ministry of Electronics and Information Technology of the Government of India, EkStep Foundation and Nilekani Philanthropies for their generous grant. We also express our sincere thanks to the Centre for Development of Advanced Computing, Pune (CDAC Pune), for providing access to their PARAM-Siddhi supercomputer, which was instrumental in model training and preparation of the benchmark. Our heartfelt appreciation goes out to all the participants involved in the recording process for data collection. Special thanks to the Hindi voice artists Rishi Dalal, Vansh Bharat Jain and Afifa Anjum, the Tamil voice artists Sam Imayavan, Clifford B. and Muthathal Subramanian and the language experts Prachi D. and Suganthi V. 

\bibliographystyle{IEEEtran}
\bibliography{refs}

\end{document}